# ELSA: A Throughput-Optimized Design of an LSTM Accelerator for Energy-Constrained Devices

ELHAM AZARI and SARMA VRUDHULA, Arizona State University

The next significant step in the evolution and proliferation of artificial intelligence technology will be the integration of neural network (NN) models within embedded and mobile systems. This calls for the design of compact, energy efficient NN models in silicon. In this paper, we present a scalable ASIC design of an LSTM accelerator named ELSA, that is suitable for energy-constrained devices. It includes several architectural innovations to achieve small area and high energy efficiency. To reduce the area and power consumption of the overall design, the compute-intensive units of ELSA employ approximate multiplications and still achieve high performance and accuracy. The performance is further improved through efficient synchronization of the elastic pipeline stages to maximize the utilization. The paper also includes a performance model of ELSA, as a function of the hidden nodes and time steps, permitting its use for the evaluation of any LSTM application. ELSA was implemented in RTL and was synthesized and placed and routed in 65nm technology. Its functionality is demonstrated for language modeling – a common application of LSTM. ELSA is compared against a baseline implementation of an LSTM accelerator with standard functional units and without any of the architectural innovations of ELSA. The paper demonstrates that ELSA can achieve significant improvements in power, area and energy-efficiency when compared to the baseline design and several ASIC implementations reported in the literature, making it suitable for use in embedded systems and real-time applications.

## 1 INTRODUCTION

**Motivation:** Among the numerous neural network (NN) models, recurrent neural networks (RNN), which are distinguished by the presence of feedback connections, have been shown to be much better suited than feed-forward NNs (e.g. CNN) for many sequence labeling tasks in the field of machine learning (ML) [10, 32]. RNNs are designed to capture the temporal dependencies within data sequences, and have been shown to learn the long-term trends and patterns inherent in sequences. To alleviate the *vanishing gradient* problem [14] in standard RNNs, and be able to learn the patterns over a larger number of time steps, more advanced RNN models, such as Gated

Authors' address: Elham Azari, eazari@asu.edu; Sarma Vrudhula, vrudhula@asu.edu, Arizona State University, Tempe, Arizona, 85281.

Recurrent Unit (GRU) [4] and Long-Short Term Memory (LSTM) [15] have been developed. The LSTM model, which is the focus of this work, has been shown to be highly robust and accurate for many applications involving time series data, including natural language processing [6] and video analysis [29]. It is now used in virtual assistant user interfaces such as *Apple Siri*, *Amazon Alexa* and *Google Assistant*. Such applications are typically launched on mobile devices, but due to their compute-intensive nature, they are executed on cloud servers. With the emergence of *Internet of Things* (IoT) and the further proliferation of mobile devices, this approach will not be scalable, and hence there is a need to move some or all of the NN computations to (energy-constrained, performance-limited) mobile devices. This poses difficult challenges associated with simultaneously achieving high energy-efficiency and high throughput. These challenges are due to the recursive structure of the LSTM model and the compute-intensive operations on very large dimensional data as well as the high memory-bandwidth requirement for computing on a large number of parameters. The goal of this paper is to achieve high energy efficiency by employing low power and compact computation units and aggressively maximizing the overall throughput.

**Literature Review:** Implementations of LSTM on CPU-GPU architectures [7, 17, 30] are not suitable for mobile devices because of their high power consumption. Although, implementations of LSTM on FPGAs [2, 3, 8, 9, 11–13, 20, 21, 24–26, 33], have been shown to be much more energy-efficient than GPUs, their power consumption is still too high (usually more than 10W) for energy-constrained systems. This has motivated the design of ASICs for LSTMs [5, 23, 27, 34].

TPU [23] describes a hardware accelerator for inferencing at cloud-scale, specialized for CNNs, Multi-Layer Perceptrons and LSTMs. Despite achieving high throughput (2.8-3.7 TOPS), it consumes up to 40W (not suitable for edge devices) and has low utilization when using LSTMs.

ASIC implementations described in [5, 27, 34] report power consumptions in the tens of milli-watts, while achieving sufficiently high throughput. Wang et al. [34] present a memory efficient ASIC design for on-line training and classification and demonstrate its functionality for language modeling and speech recognition. To eliminate off-chip memory use, the model parameters are reduced by using circulant matrices along with a compression technique. The reported results are only at synthesis level. DNPU [27] is a reconfigurable CNN-RNN processor with the CNN being its major component and the LSTM as its secondary unit. Quantization and table-based multiplication techniques result in significant reduction in on-chip memory storage. However, this limits its peak performance by requiring the use of external memory even for models with small number of parameters. CHIPMUNK [5] is a scalable LSTM accelerator that is designed to handle applications that operate on large datasets. It achieves significant reduction in the memory transfer overhead by allowing the interconnection of multiple LSTM units in a systolic array structure. However, the power consumption of the *systolic array structure* is too high at an application level and not suitable for energy-constrained edge-devices.

There exists CNN ASIC implementations that employ efficient multipliers for performing the convolution operations [1, 18]. Albericio et al. [1] present a massively data-parallel architecture for CNNs that eliminates most of the ineffectual computations by using a serial-parallel shift-and-add multiplication (Pragmatic unit). The Pragmatic unit as well as its previous bit-serial version (Stripes [18]) are designed to efficiently perform the *inner products* in the convolution layers and also skip the zero bits in the activations. Albericio et al. [1] report 92% zero bits in the activation values of CNNs. One of the reasons is due to the rectified linear (ReLU) activation function that converts negative activations to zero, resulting in many zero activations and no negative activations. Although the multiplier designs in [1] and [18] lead to improvements in energy efficiency as compared to its equivalent state-of-the-art CNN accelerators, they cannot be used in the multiplication operations in an LSTM network because of two main reasons. First, there is no convolution operation involved in a typical LSTM network. Second, many of the zero bits in the activations of CNNs as shown in [1] are due to the ReLU activation function, which does not exist in a typical LSTM network. The sigmoid and tanh functions are used in an LSTM network that have different properties as compared to ReLU.

**Contributions of this work:** Existing ASIC implementations of the LSTM model are based on *conventional* architectures. This paper describes the design of an **e**nergy-efficient **LS**TM **a**ccelerator, referred to as ELSA. The overarching goal of this work is to aggressively reduce the power consumption and area of the LSTM components, and then use architectural level techniques to boost the performance. This is achieved by two main steps. First, we design and employ low power and compact computation units for the LSTM. Some of these modules use approximate calculations, which require much lower power but incur a high execution time penalty, i.e. it may take multiple clock cycles to finish one operation. Moreover, many of these modules are on the critical path which further degrade the performance. Second, to recover the throughput loss and achieve higher energy efficiency, we develop efficient scheduling techniques that include overlapping of the computations at multiple levels – from the lowest level modules up to the application. The main results of this work are summarized below.

(1) The performance of a low power approximate multiplier (AM) is significantly improved and incorporated in the compute-intensive units of ELSA. The execution time of the AM is data-dependent and the number of clock cycles required to finish a single multiplication depends on the magnitude of the multiplicand. An intricate hierarchical control with four distinct, interacting controllers are designed to efficiently synchronize the single-cycle and variable-cycle operations in ELSA.
(2) To maximize the throughput and compensate for the performance loss, elastic pipeline stages are incorporated at *three* levels. The first one is at the MAC level as these units are *internally pipelined* simultaneously. The second and third levels are at the LSTM layer (overlapping the operations at different time steps) and application, respectively.
(3) A general performance model of ELSA as a function of hidden nodes and time steps is also presented. This is to permit accurate evaluation of ELSA for any application that includes a network of LSTM layers, such as speech recognition and image captioning.

**Paper Organization:** Section 2 describes the LSTM structure and its key computations. Section 3 describes a significantly improved version of an existing approximate multiplier. It justifies the proposed multiplier and describes the design challenges it poses. Section 3 also explains the ELSA's architecture including its controllers. Section 4 explains the multi-level elastic pipelining and Section 5 includes the performance models for ELSA. Section 6 explains the application implemented for demonstrating the functionality of ELSA. The ASIC implementation results are compared with the baseline-LSTM and two existing implementations. Section 6 also demonstrates the accuracy of ELSA as compared to floating-point and exact fixed-point designs. Section 7 concludes the paper.

## 2 BACKGROUND
### 2.1 Long Short-Term Memory

Figure 1 shows a typical LSTM layer. The input is a temporal sequence $X = (x_1, x_2, \cdots, x_T)$ and the output is a sequence $h = (h_1, h_2, \cdots, h_T)$, referred to as the *hidden state*, that is generated iteratively over $T$ time steps. The memory cell ($C$) stores some part of the past history over a specific period of time. At each iteration, the *input gate* controls the fraction of the input data to be remembered and the *forget gate* determines how much of the previous history needs to be deleted from the current memory state ($C_t$). The output gate decides how much of the processed information needs to be generated as the output ($h_t$). In a sequence learning task, let $X = (x_1, x_2, \cdots, x_T)$, where $x_t$ is the input to the LSTM layer at time step $t \in [1, 2, ..., T]$. The following equations show how the output sequence $h = (h_1, h_2, \cdots, h_T)$ of a layer is generated iteratively over $T$ time steps:

$$i_t = \sigma(W_{xi}x_t + W_{hi}h_{t-1} + b_i), \tag{1}$$
$$o_t = \sigma(W_{xo}x_t + W_{ho}h_{t-1} + b_o), \tag{2}$$
$$f_t = \sigma(W_{xf}x_t + W_{hf}h_{t-1} + b_f), \tag{3}$$
$$\hat{C}_t = \tanh(W_{xc}x_t + W_{hc}h_{t-1} + b_c), \tag{4}$$
$$C_t = i_t \odot \hat{C}_t + f_t \odot C_{t-1}, \tag{5}$$
$$h_t = o_t \odot \tanh(C_t). \tag{6}$$

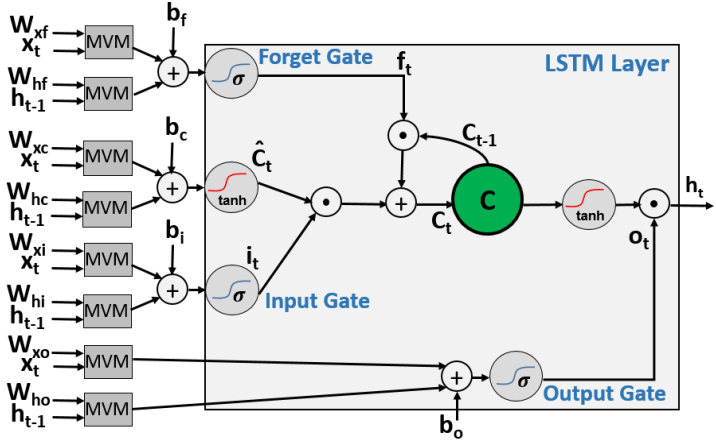

Fig. 1. The Structure of an LSTM layer. It consists of a memory cell (C), an input gate (i), an output gate (o), and a forget gate (f). MVM = matrix-vector multiplier; • = element-wise multiplier; $\sigma$, tanh = sigmoid and hyperbolic tangent activation functions.

The element-wise multiplication is indicated by $\odot$. The parameters are the bias vectors ($b$s) and the weight matrices ($W$s) which are tuned during model training. $\hat{C}_t$ is the new candidate memory which contains the extracted information from the input. The non-linear activation functions, $\sigma \in (0, 1)$ and $tanh \in (-1, 1)$, are defined in Equations 7 and 8.

$$\sigma(x) = \frac{1}{1 + e^{-x}} \tag{7}$$
$$tanh(x) = \frac{e^x - e^{-x}}{e^x + e^{-x}} \tag{8}$$

The main challenges in the design of an LSTM architecture is the large number of matrix-vector multiplications (MVMs) involving large dimensional vectors, the element-wise multiplications (EMs) and the data movements from/to the memory.

## 3 ARCHITECTURE OF ELSA

### 3.1 Approximate Multiplier (AM)

The advantages of the approximate multiplier (AM) design in [28] are its low logic complexity and reduced power consumption. The AM generates an *approximate* (but sufficiently accurate) product.

The inputs and outputs of the AM are represented as signed, fixed-point fractions, i.e., in a binary fraction $X = x_{n-1}.x_{n-2}x_{n-3}...x_0$, the sign bit is $x_{n-1}$ and the fraction is $x_{n-2}x_{n-3}...x_0$. Let $\mathcal{N}$ denote the numerator of a fraction. For example, in a 4 bit number $X = 1.010$, $\mathcal{N}(X) = -6$ and its decimal value is $-6/8$. Let $X$ and $W \in [-1, 1)$ be the inputs of the AM in an $n$ bit multiplication. The exact product is $XW$. The AM produces an n-bit $Z \approx XW$. The main component of the AM is an FSM (with $2^n$ states) that generates a specific bit-stream $\{S\}$. The generation of this bit-stream depends on the value of one of the operands, say $X$ and its length is $|\mathcal{N}(W)|$. Specifically, in $\{S\}$, $X_{n-i}$ appears at cycle $2^{i-1}$, and then after every $2^i$ cycles, for $i \in [1, n]$. The main property of $\{S\}$ is that the difference in the number of ones and zeros is an integer approximation to $XW$. The theoretical upper-bound on the approximation error is $n/2^{n+1}$, but has been empirically shown to be far smaller, approaching the precision of floating point for $n \geq 8$ [28]. The structure of a 4-bit signed AM along with an example is illustrated in Figure 2. In this example, $X = 5/8$, $W = 6/8$ and $n = 4$. The FSM consists of $2^4$ states. The FSM-MUX combination generates the bit-stream $S = \{\overline{x_3}\,x_2\,\overline{x_3}\,x_1\,\overline{x_3}\,x_2\}$ at cycles $\{c_1\,c_2\,c_3\,c_4\,c_5\,c_6\}$, respectively. The up-down counter counts up as it receives one and counts down when it receives a zero and generates the product Z. The down-counter stops the AM after $|\mathcal{N}(W)|$ cycles, which is 6 (i.e. $(6/8) \times 2^3$), in this example. The XOR gate receives the bit-stream $\{S\}$ and the MSB of W ($w_3$), and generates the input of the up-down counter in sequence. Then, the output of the AM is produced by the up-down counter, which is 0.5. The exact result is 0.46, as shown in the table.

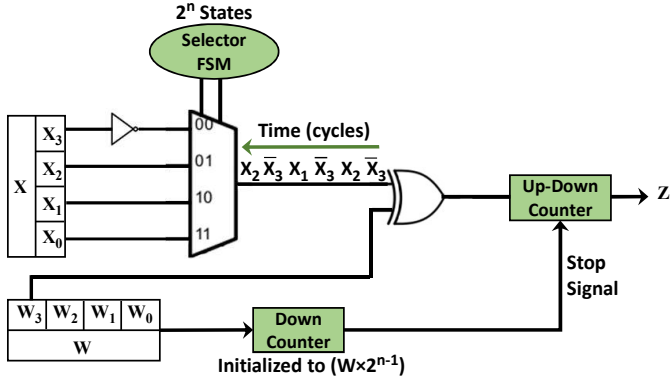

| X | W | Input of the Up-Down Counter | Output of the AM | Exact Result |
|---|---|---|---|---|
| $x_3\,x_2\,x_1\,x_0$<br>0. 1 0 1 | $w_3\,w_2\,w_1\,w_0$<br>0. 1 1 0 | $c_6\,c_5\,c_4\,c_3\,c_2\,c_1$<br>1 1 0 1 1 1 | $\dfrac{4}{8} = 0.50$ | $\dfrac{30}{64} = 0.46$ |

| Init. Value of the Down Counter | Init. Value of the Up-Down Counter |
|---|---|
| 6 | 0 |

Fig. 2. Structure of a 4-bit signed AM. $X$ and $W \in [-1,1)$ are the inputs and $Z$ is the product. The sign bits are $x_3$ and $w_3$.

### 3.2 Extension to AM For a Faster Execution

The AM is a compact design with low power consumption, however, its execution time is high as compared to the fixed-point exact multiplier. Hence, the design of the original AM is modified to improve its execution time by 2X with negligible logic overhead (< 0.03%), as shown in Figure 3.

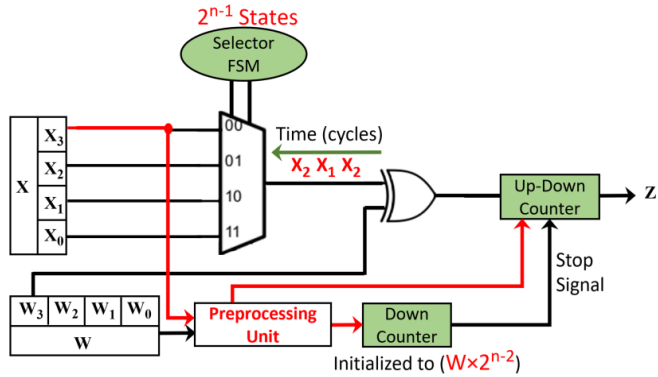

| X | W | Input of the Up-Down Counter | Output of the AM | Exact Result |
|---|---|---|---|---|
| $x_3\ x_2\ x_1\ x_0$ | $w_3\ w_2\ w_1\ w_0$ | $c_3\ c_2\ c_1$ | $\dfrac{4}{8} = 0.50$ | $\dfrac{30}{64} = 0.46$ |
| 0.  1   0   1 | 0.  1   1   0 | 1   0   1 | | |

| Init. Value of the Down Counter | Init. Value of the Up-Down Counter |
|---|---|
| 3 | 3 |

Fig. 3. The improved version of the AM. The modified parts are shown in red. The number of states in the FSM is reduced by half and the down counter is initialized to half of its value as compared to the one in Figure 2. The preprocessing unit sets the initial value of the up-down counter to 3 (i.e, $\overline{X_3}$ is one and it appears three times in the bit-stream in Figure 2). Hence, the initial value of the down-counter is set to 3, half of its original value.

This is achieved by adding a small *preprocessing unit* to extract the FSM patterns for the MSB of the first operand and initializing the up-down counter by the computed value. As the FSM in the original design selects the MSB every two cycles, this modification leads to decreasing the latency of the original AM by 50%. The preprocessing unit consists of an *inverter*, a *shifter* and an *XOR* gate. This unit receives the MSB of X (i.e. $\overline{X_3}$ in this example) and $W$ as its inputs, and generates two outputs as the initial values of the down-counter and the up-down counter. It shifts the value of W to the right by one and sets it as the initial value of the down-counter. The same operation is performed to set the initial value of the up-down counter, except that the sign of the computed value needs to be specified. The sign is determined based on the result of $(\overline{X_3} \oplus W_3)$ which is computed once in the preprocessing unit. If it is a one, the sign is positive, otherwise it is negative. For the example shown in Figure 2, the FSM-MUX in this AM generates the bit-stream $S = \{x_2\ x_1\ x_2\}$ at cycles $\{c_1\ c_2\ c_3\}$, respectively. This results in saving 3 (50%) cycles as compared to the one shown in Figure 2. The multipliers in ELSA employ this accelerated AM to achieve higher throughput, while maintaining low area and power consumption.

### 3.3 Comparison with an Exact Multiplier

Employing AMs to perform the compute-intensive operations (i.e. MVM) can result in significant savings in both area utilization and power consumption. To explore this, the AM (labeled AM-MAC) and an exact fixed-point multiplier (labeled Exact-MAC) were designed and compared when used in MAC units. Each of these MACs consist of 100 individual multipliers and adders to perform 100 MAC operations in parallel. These units were synthesized using Cadence® GENUS running at 200MHz, for various bit widths ranging from 8 to 16 bits. Figure 4 shows the improvement in power and cell area of AM-MAC as compared to the Exact-MAC. Each plot also shows the accuracy of the AM-MAC as compared to the Exact-MAC for various bit widths. As the bit precision increases, the accuracy of the AM-MAC improves from 97.8% to 99.9%, and the maximum savings in the power consumption and cell area reaches 79.49% and 63.30%, respectively. **Note:** Delay comparison of these units in isolation is not meaningful as the AM requires variable number of cycles (i.e. data-dependent) for a single multiplication. Delay comparison of an LSTM network for an application is more meaningful and is described and quantified in Section 6.

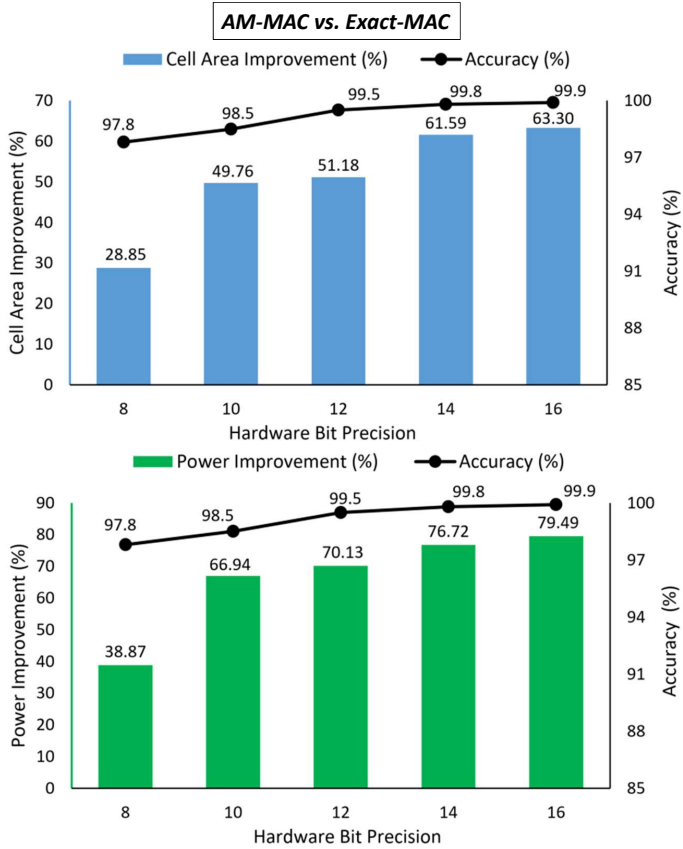

Fig. 4. The cell area improvement (top) and power improvement (bottom) of AM-MAC as compared to the Exact-MAC for different hardware bit precision. Each plot demonstrates the accuracy of the AM-MAC for different bit precision as well.

### 3.4 Hardware Challenges and Design Decisions

Although employing AM leads to substantial reduction in area and power consumption, this variable-cycle multiplier poses a number of design challenges. One is the increased latency of the MVM and EM units, both of which lie on the critical path. ELSA's modification of the AM includes elimination of one state of the FSM, and results in improving the AM's performance by 2X. This in itself is not enough. Hence, ELSA's design maximally overlaps the execution of the MVMs with other computation units and over multiple time steps, resulting in a multi-level pipelined design.

In addition, the control unit is organized as a two-level hierarchy to efficiently synchronize the AM units and overlap their computations to practically eliminate the waiting time (e.g. arising from being a variable-cycle multiplier) and hide their latency. Finally the potential loss of accuracy due to the presence of feedback and use of AMs is addressed by experimentally evaluating the optimal bit precision for the overall design. The optimal bit precision of ELSA is evaluated by comparing its accuracy with two corresponding LSTM designs. The first one is the software implementation with floating-point calculations, and the second one is an LSTM design with *exact* fixed-point multiplications. This is performed for the following reasons:

(1) to explore the impact of using the AMs in ELSA on error propagation through the LSTM for different bit precision and to investigate whether the error accumulates in the hidden and memory states over various time-steps. This is performed by measuring the mean squared error between the hidden states/memory states of ELSA and the floating point implementation.
(2) to evaluate the best hardware bit-precision for ELSA that is a good trade-off between its accuracy and its hardware design metrics (i.e. power, area, performance). This is performed by calculating the classification accuracy of ELSA and comparing it with its corresponding exact fixed-point implementation.

## 3.5 System Overview

The top-level block diagram of ELSA is shown in Figure 5. It includes the computation units in LSTM as well as the controllers that synchronize them. The computation units include: 1) the MVM modules to perform the matrix-vector multiplications in an LSTM layer in parallel. 2) The ternary adders to perform the addition on the outputs of the MVMs and the bias vectors. 3) the non-linear activation functions, i.e., sigmoid and tanh. 4) EMA module to compute the elements of the memory state. 5) EM module to compute the elements of the hidden state. The control of the AM units is performed by a top controller in coordination with three distinct mini controllers- MVM-C, EM-C and EMA-C. The reason to include mini controllers is that the computation modules that employ the AM units (i.e. MVM, EM, EMA), involve variable-cycle operations and hence require synchronization with other single-cycle operations. Moreover, these units have to execute in parallel to maximize throughput. The required network parameters are loaded into the SRAMs, and the data transfer for fetching/storing the parameters from memory is controlled by the controller units. The intermediate results of the computation units are written into the buffers to reduce the SRAM access time. Thus, the SRAMs are only accessed for fetching the parameters and storing the computed values for the hidden and memory-states. The components of ELSA as well as the multi-level elastic pipelining technique are explained in details in the following sections.

## 3.6 Main Computation Units

**MVM Module:** The MVM module is a compact combination of the AM units that receives a matrix $X_{n \times m}$ and a vector $Y_{m \times 1}$ as inputs. There are totally $n$ AM units in an MVM module that all share the same FSM and down-counter, thereby making the module compact. This unit is *internally pipelined* with $m$ pipeline stages. The parallel matrix vector multiplication in the MVM module is performed by multiplying one column of matrix X with one element of vector Y at a time. To store the MAC results, the up-down counter performs as an accumulator and its bit-width is increased by a few bits to preserve the precision. In the example shown in Figure 6, at time $t_1$, the first column-scalar multiplication is performed on column $[x_{11}, x_{21}, x_{31}]^T$ and scalar $y_1$. The latency of these multiplications which execute in parallel is determined by $y_1$, and the first partial results are accumulated in the up-down counters. Without resetting the up-down counters, this process is repeated until time $t_4$, at which the last column-scalar multiplication (i.e. $[x_{14}, x_{24}, x_{34}]^T \times y_4$) is computed and the final output vector $Z_{3\times1}$ is generated. As shown in Figure 6, the difference between the start and end times of the operations are not necessarily equal due to their variable cycle execution.

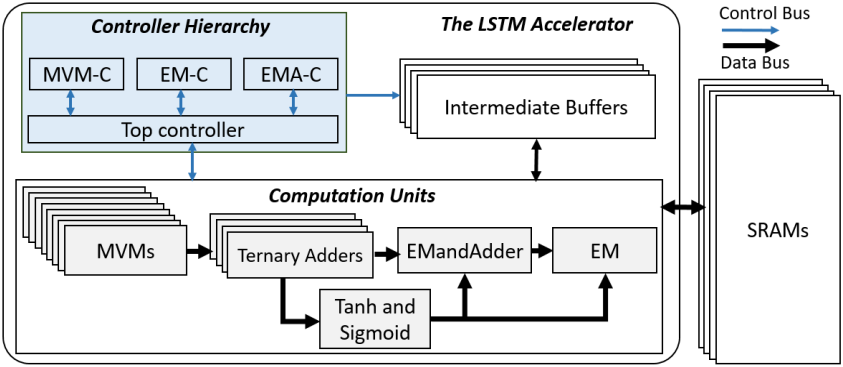

Fig. 5. The block diagram of ELSA that consists of the computation units and a hierarchy of control units.

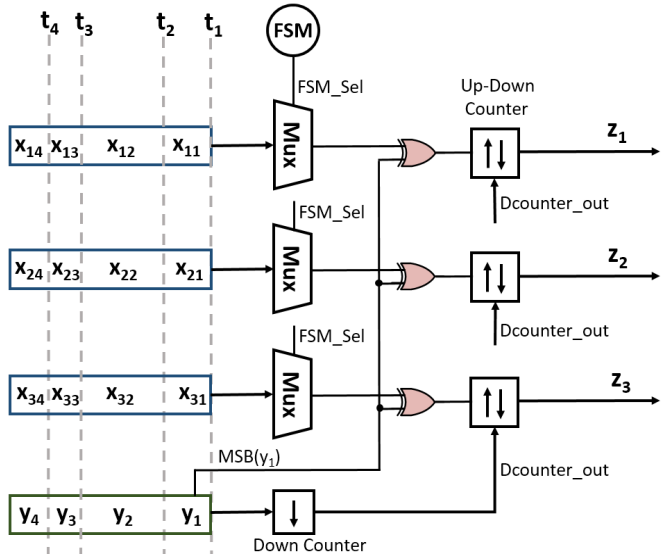

Fig. 6. An example of the MVM module, in which it receives $X_{3\times4}$ and $Y_{4\times1}$ as its inputs and generates the output vector $Z_{3\times1}$.

**EM and EMA modules:** The Element-wise Multiplier (EM) and Element-wise Multiplier and Adder (EMA) modules employ the accelerated AM shown in Section 3.2 to compute the components of the $h$ and $C$ vectors, respectively.

**Sigmoid and Tanh Modules:** The non-linear activation functions can be implemented in hardware using polynomial approximations [22], look-up tables, or CORDIC algorithms [16]. These implementations utilize large area and consume high power. Therefore, $\sigma$ and $tanh$ in ELSA are implemented as piece-wise linear functions [34], as shown in Table 1, resulting in a more compact and lower power design.

Table 1. Piece-wise linear activation functions [34].

$$\mathbf{HSig(x)} = \begin{cases} +1 & x > 2 \\ \frac{x}{4} + 0.5 & otherwise \\ 0 & x \leq -2 \end{cases} \quad \mathbf{HTanh(x)} = \begin{cases} +1 & x > 1 \\ x & otherwise \\ -1 & x \leq -1 \end{cases}$$

## 3.7 Controller Units

Figure 7 shows the control flow graph (CFG) of the top-level controller (Top-C). It consists of three mini controllers – *MVM-C*, *EM-C* and *EMA-C*. The computation modules that involve variable-cycle operations (i.e. MVM, EM, EMA) require synchronization with other single-cycle operations (e.g. adders). Moreover, these units have to execute in parallel to maximize throughput. This cannot be accomplished by Top-C alone. The mini controllers are designed to individually control the AM-based units.

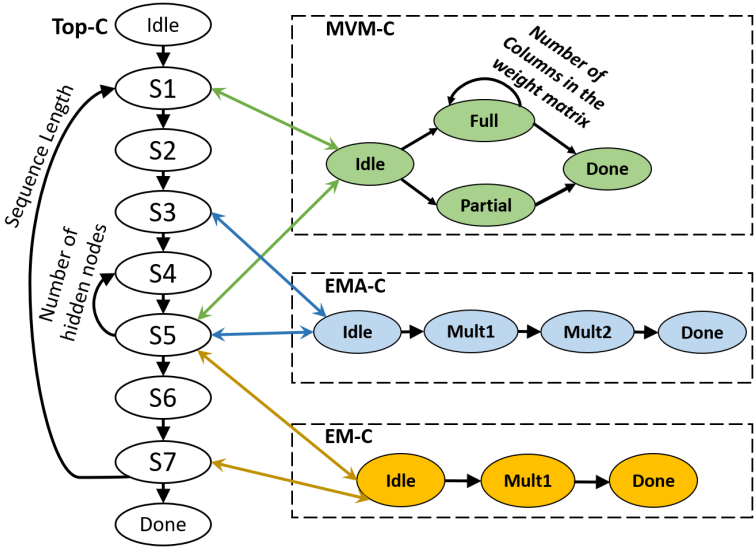

Fig. 7. The controller hierarchy that consists of a top controller (Top-C) and three mini controllers- MVM-C, EMA-C and EM-C.

**Top Controller (Top-C):** This is responsible for synchronizing the AM-based modules with other single-cycle units and enabling parallel executions. As shown in Figure 7, it consists of 7 different states, where states $S1$, $S3$, $S5$ and $S7$ activate the MVM, EM and EMA modules. For example, when Top-C is in $S1$, the control token is passed to the MVM-C to start the MVM operations. The MVM-C operates on one set of data for multiple cycles and generates a complete detection signal that sends the control back to the Top-C. This is the case for all the Top-C states that call the mini controllers.
**MVM mini-Controller (MVM-C):** This activates the MVM modules and consists of two major states – *partial* and *full*. The *full* state is responsible for operating on all the columns of the matrix iteratively to compute the complete results. This state is used to generate the initial data for the pipelining flow. The *partial* state only operates on one column-scalar multiplication to generate one partial result. This state is designed to overlap its computation with the *EM* and *EMA* units which are active in $S5$ of Top-C.

Table 2. This shows the control flow and the data computation in the proposed pipelining method. $T$ is the total number of time steps. $n$ is the total number of hidden nodes, and $j$ denotes the $j_{th}$ component of its corresponding vector. The output of the stage operations as well as the mode of operation for the MVMs are specified below. The pipeline stages shown in columns are executed in parallel and the stages shown in rows are performed sequentially.

| Multiple Cycles | One Cycle | Multiple Cycles | One Cycle | Multiple Cycles | One Cycle | Multiple Cycles |
|---|---|---|---|---|---|---|
| Stage 1 Ops (t) *full* | Stage 2 Ops (t) $f_j(t), \hat{C}_j(t), i_j(t)$ | Stage 3 Ops (t) $C_j(t)$ | Stage 2 Ops (t) $f_{j+1}(t), \hat{C}_{j+1}(t), i_{j+1}(t)$ Stage 4 Ops (t) $o_j(t)$ Stage 5 Ops (t) $tanhout_j(t)$ | Stage 6 Ops (t) → Stage 1 Ops (t+1) $h_j(t)$, *partial* Stage 3 Ops (t) $C_{j+1}(t)$ | Stage 5 Ops (t) $tanhout_n(t)$ | Stage 6 Ops (t) $h_n(t)$ |
| Controller State 1 | Controller State 2 | Controller State 3 | Controller State 4 | Controller State 5 | Controller State 6 | Controller State 7 |

**EM mini-Controller (EM-C):** The EM-C consists of one multiplication state to control the EM computation units. Once the operation is done, it sends the control back to Top-C, which then activates the MVM-C for overlapping the data computation in time steps $t + 1$ and $t$.

**EMA mini-Controller (EMA-C):** The EMA-C includes two consecutive multiplication states (i.e. Mult1 and Mult2) to activate the EMandAdder unit for generating one component of the memory state vector at each iteration.

## 4 MULTI-LEVEL ELASTIC PIPELINING

Some of the computation units in an LSTM network have data dependencies among themselves. These have to be executed sequentially, while others can execute in parallel. Although a non-pipelined version is straightforward, the throughput would be unacceptably low. Pipelining is essential and ELSA incorporates pipelining at *three* levels, involving variable-cycle multipliers, various computation units within the LSTM layer, and across multiple time-steps.

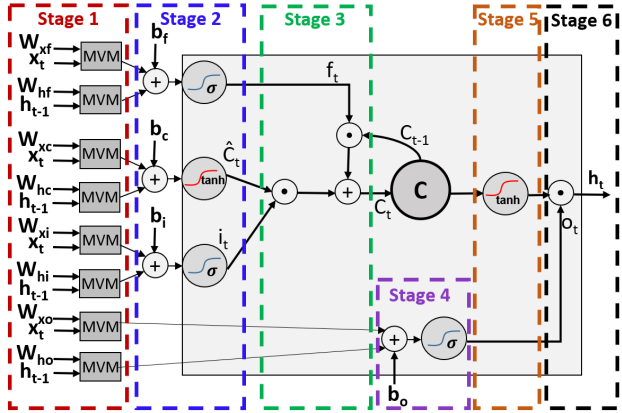

Fig. 8. The six pipeline stages in the LSTM layer. $Stage-1$: Eight parallel MVMs; $Stage-2$: Three activation functions and ternary adders; $Stage-3$: two consecutive multiplications and an adder; $Stage-4$: a sigmoid function and a ternary adder; $Stage-5$: one tanh function; $Stage-6$: one element-wise multiplication.

ELSA consists of six elastic pipeline stages as shown in Figure 8. The latency of some of these stages are multi-cycle and conventional pipelining methods are not efficient enough to maximize the throughput of this design. Table 2 shows the control flow of the pipelining method along with the data computations done in each controller state. The overlapping of the computation units starts in controller state 4 where the operations in pipeline stages 2, 4 and 5 at time step $t$ are performed in parallel.

In controller state 5, the operations in stage-6 (time step $t$) and stage-1 (time step $t + 1$) are overlapped with two consecutive multiplications in stage-3 (time step $t$). Since the stage-3 operations is independent of the ones in Stages 6 and 1, they can be executed in parallel. It is worth mentioning that with the proposed scheme, the MVMs are almost completely overlapped with other units, as are the memory accesses, resulting in near maximum resource and memory utilization. All the intermediate results are written into the buffers so the SRAMs are only accessed for fetching the parameters and writing back the computed values for the hidden-state (h) and memory-State (C). These result in substantial reduction in the overall design latency as well as maximizing the throughput. These are quantified in Section 5 and Section 6.

## 5 PERFORMANCE MODELING FOR ELSA

This section presents a general model for the execution time of ELSA as a function of hidden nodes and time steps. This is to permit accurate evaluation of ELSA for any application that includes a network of LSTM layers, e.g., speech recognition, image captioning, etc. A similar performance model for the non-pipelined version of ELSA is also constructed to quantify the improvements due to the pipelining strategy employed in ELSA.

Let $X = (X_1, X_2, \cdots, X_T)$ and $H = (H_1, H_2, \cdots, H_T)$, where $X_t$ and $H_t$ are the input and output of ELSA at time step $t \in [1, 2, \cdots, T]$, respectively. In an LSTM layer with N hidden nodes, $X_t = [x_t^1, x_t^2, ..., x_t^N]$ and $H_t = [h_t^1, h_t^2, ..., h_t^N]$.

As discussed in Section 4, each controller state may contain a single pipeline stage (e.g. controller state 2) or multiple pipeline stages (e.g. controller state 4). The execution time (D) of each controller state (CS) is denoted by $D_{CS_i}$, for $i \in [1, 2, ..., 7]$. The execution time is expressed in number of clock cycles. The operations performed in $CS_2$, $CS_4$ and $CS_6$ are single cycle operations whereas those in $CS_1$, $CS_3$, $CS_5$ and $CS_7$ are multi-cycle operations, whose latency is data-dependent and determined during run-time. The execution time of these operations is expressed in terms of the magnitude of their multiplicands (e.g. $\|x_t^j\|$ in stage 1 of Figure 8, where $t$ denotes the time step and $j$ is the $j_{th}$ component of the $X_t$ vector). This is because of the AM units, in which the multiplicands determine the execution time in number of clock cycles. In all the equations, $j \in [1, 2, ..., N]$, $t \in [2, ..., T]$ and $i$, $o$ and $f$ correspond to the input, output and forget gates, respectively. Note that the following equations can be directly derived from Table 2 and Figure 8.

### 5.1 Pipelined Design

The delay equations for ELSA with multi-level pipelining are shown in Equations 9-15, after the initial data is produced to flow through the pipeline stages (i.e t>2). The quantities in the equations correspond to the variables in Table 2. For example, since the MVM modules execute in parallel and X and H determine the execution time of these operations, $D_{CS_1}$ in Equation 9 is the maximum value of each component of these vectors.

$$D_{CS_1}(t) = \max(\|x_{t+1}^N\|, \|h_t^N\|), \tag{9}$$

$$D_{CS_3}(t) = \frac{\|i_t^1\| + \|f_t^1\|}{2}, \tag{10}$$

$$D_{CS_5}(j, t) = \max\left(\frac{\|o_t^j\|}{2} + \max_{t \neq T}\left(\|x_{t+1}^j\|, \|h_t^j\|\right), \frac{\|i_t^{j+1}\| + \|f_t^{j+1}\|}{2}\right), \quad (11)$$

$$D_{CS_7}(t) = \frac{\|o_N^t\|}{2}, \quad (12)$$

$$D_{CS_2} = D_{CS_4} = D_{CS_6} = 1. \quad (13)$$

The total execution time of ELSA with pipelining ($D_{Total}^p(j, t)$), which is a function of hidden nodes and time steps is shown in Equation 14 and is simplified in Equation 15.

$$\begin{aligned} D_{Total}^p(j, t) &= \sum_{t=2}^{T}(D_{CS_1}(t) + D_{CS_3}(t) + 2 + D_{CS_7}(t)) \\ &+ \sum_{t=2}^{T}\sum_{j=1}^{N-1}(D_{CS_5}(j, t) + 1) \end{aligned} \quad (14)$$

$$\begin{aligned} D_{Total}^p(j, t) &= \sum_{t=2}^{T}(D_{CS_1}(t) + D_{CS_3}(t) + D_{CS_7}(t)) \\ &+ \sum_{t=2}^{T}\sum_{j=1}^{N-1}(D_{CS_5}(j, t)) + T + N(T-1) - 1 \end{aligned} \quad (15)$$

### 5.2 Non-Pipelined Design

The delay equations for the non-pipelined design are shown in Equations 16-20. Note that the same units and structure are used for both the designs. The only difference between these two designs are the way the operations are executed. In the non-pipelined version, the stages shown in Figure 8 execute in sequence. Hence, the execution time is expressed in terms of the pipeline stages, and does not correspond to the control sequence shown in Figure 7.

$$D_{stage_1}(j, t) = \sum_{j=1}^{N} \max(\|x_t^j\|, \|h_{t-1}^j\|), \quad (16)$$

$$D_{stage_3}(j, t) = \frac{\|i_t^j\| + \|f_t^j\|}{2}, \quad (17)$$

$$D_{stage_6}(j, t) = \frac{\|o_t^j\|}{2}, \quad (18)$$

$$D_{stage_2} = D_{stage_4} = D_{stage_5} = 1. \quad (19)$$

The total execution time of the non-pipelined design, which is denoted by $D_{Total}^{np}(j, t)$, is shown in Equation 20.

$$\begin{aligned} D_{Total}^{np}(j, t) &= \sum_{t=2}^{T}(D_{stage_1}(j, t)) + 3N(T-1) \\ &+ \sum_{t=2}^{T}\sum_{j=1}^{N}(D_{stage_3}(j, t) + D_{stage_6}(j, t)) \end{aligned} \quad (20)$$

To compute the impact of the pipelining method on the overall execution time of ELSA, equations 15 and 20 were evaluated and compared for different bit precision, hidden nodes and time steps. These are shown in Table 3. Thus, a total of 27 configurations were evaluated. Based on empirical data, the pipelining alone achieves 1.62X improvement in performance on average as compared to the non-pipelined design. The speedup achieved for each configuration was close to 1.62X, so only the average is reported.

Table 3. The average speed-up achieved by the pipelining method over the non-pipelined design for 27 different LSTM configurations. This was computed by evaluating Equations 15 and 20. The minimum and maximum speedups were 1.58X and 1.65X, respectively.

| 27 Different LSTM Configurations | | | | Average Speedup (X) over 27 Configurations |
|---|---|---|---|---|
| *Bit Precision* | 8 | 12 | 16 | |
| *Hidden Nodes* | 64 | 128 | 256 | **1.62X** |
| *Time Steps* | 10 | 100 | 1000 | |

## 6 EXPERIMENTAL RESULTS

### 6.1 Application

The functionality of ELSA is demonstrated for character-level Language Modeling (LM), which is one of the most widely used tasks in natural language processing [31]. LM predicts the next character given a sequence of previous character inputs. It generates text, character by character that captures the *style and structure* of the training dataset. The text that is produced *looks like* the original training set. The LM used in this paper for training was written using a scientific computing framework referred to as Torch [19]. It consists of two 128-hidden node LSTM layers. For the evaluation in this paper, the model was trained on a subset of Shakespeare's works by setting the batch size, training sequence and the learning rate to their default values of 50, 50 and 0.002, respectively [19]. The network architecture is shown in Figure 9. The input to this model is a character formed in a one-hot vector of size 65 (i.e. the number of characters used in this model, which includes the lower and uppercase letters with some special characters). The first LSTM layer receives this input and generates a hidden vector of size 128, which is fed as input to the second LSTM layer. Similarly, the second LSTM layer generates the final 128-node hidden vector and passes the output to the fully connected (FC) and the softmax layer. The final output is a 65-node vector whose components represent the likelihood of that corresponding character being the output.

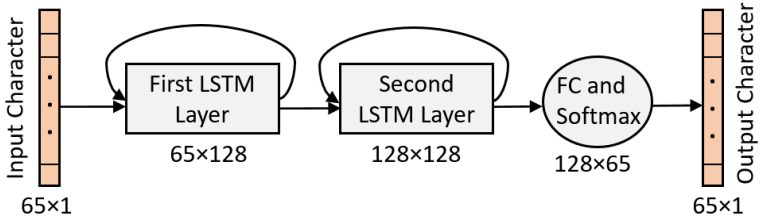

Fig. 9. The network structure of LM that consists of two LSTM layers with 128 hidden-nodes followed by FC and softmax layers.

## 6.2 ASIC Implementation of ELSA

ELSA's design was specified in RTL, synthesized and placed and routed (using Cadence® tools) in 65nm CMOS technology achieving a peak frequency of 322 MHz. ELSA's RTL design, including the controllers, is fully parametrized and can adapt to any LSTM network topology. Hence, there is no need to do the pipelining again as the controllers automatically accommodate the change. In addition, no design effort is required for varying the bit precision and modifying the size of the hidden nodes for a given application. Figure 10 shows the physical layout of ELSA's design in 65nm and the characteristics of the ASIC implementation. ELSA has sufficiently small cell area and low power, making it suitable for use in embedded systems. Moreover, the efficient scheduling and pipelining techniques led to a design with high peak performance making ELSA also suitable for real-time applications.

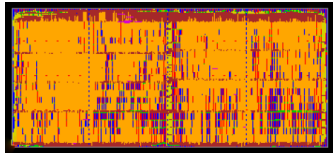

|  | **ELSA** |
|---|---|
| **Core Voltage (V)** | 1.1 |
| **Number of MACs** | 772 |
| **Precision (bit)** | 8-11 |
| **Frequency (MHz)** | 322 |
| **Total Cell Area (mm$^2$)** | SRAM area: 2.22  LSTMCell area: 0.4 |
| **On-Chip Memory (KB)** | 106 |
| **Peak Performance (GOP/s)** | 27 |
| **Power (mW)** | 20.4 |
| **Energy-Efficiency (GOP/s/mW)** | 1.32 |

Fig. 10. The physical layout of ELSA's design in 65nm CMOS technology (top) and the ASIC's implementation results (bottom).

ELSA uses an 8-bit fixed-point representation (explained in Section 6.3) with the intermediate results extended to 11 bits to preserve the precision. The SRAMs incorporated in ELSA were provided by the 65nm library supplier. Unfortunately, the available SRAMs were larger than necessary and hence their area and power numbers shown in Figure 10 should be considered as pessimistic, by as much as ∼6%. The SRAM area of ELSA is approximately 3X larger than its logic area. Clock gating of the computation units and the mini-controllers and the use of sleep modes for the SRAMs were employed to further reduce the power consumption. Because of the variable-cycle pipeline stages, ELSA's design greatly benefits from clock-gating. The greatest reduction in power was achieved when the computation units in a multi-cycle pipeline stage were maximally utilized. Hence, all the other idle units were clock-gated for several cycles.

Figure 11 shows the power and area breakdown of ELSA's components, including the SRAMs. The power consumption was measured using data activity information (*.vcd) obtained from the testbench by simulating the design after placement and routing. As expected, the SRAMs consume the most power. Among the submodules, the controllers contribute the least to the power consumption and the MVMs consume the most as there are 772 MACs in this design. There is substantial difference between the area utilization of the SRAMs and all the other components. Although there are 772 MACs in this design, the MVMs constitute to only 10.66% of the total area.

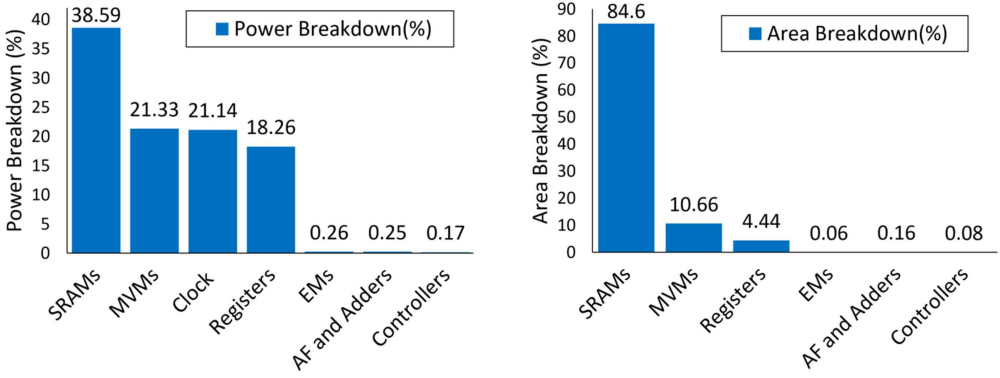

Fig. 11. Power (left) and Area (right) breakdown of ELSA's components including the SRAMs. AF stands for activation functions.

6.2.1 **Comparison with the Baseline-LSTM**. The LSTM network was also designed with 8-bit exact fixed-point multipliers and is referred to as the Baseline-LSTM. This is functionally equivalent to ELSA except that all the AM units were replaced with the exact multipliers. These multipliers were optimally synthesized by the Cadence tools (i.e. Genus) based on the clock frequency constraint. This is automatically generated by Genus to meet the timing constraints corresponding to the given clock frequency. The Baseline-LSTM was also specified in RTL and synthesized and placed and routed in 65nm technology. The ASIC implementation results of ELSA are compared with the Baseline-LSTM and the normalized results are shown in Figures 12 and 13. In Figure 12, ELSA and the Baseline-LSTM were run at the same clock frequency (the peak frequency of the Baseline-LSTM). The energy efficiency ($GOP/s/mW$) and area efficiency ($GOP/s/mm^2$) of ELSA exceeds that of the Baseline-LSTM by 1.2X. The cell area and power consumption of ELSA are also lower (0.3X), but the peak performance of the Baseline-LSTM is higher by 3.3X. This is to be expected as the operations in the Baseline-LSTM are single cycle operations and the Baseline-LSTM was run at its highest clock frequency, unlike ELSA's.

For a thorough comparison, both designs were also run at their individual maximum achievable clock frequencies. The results are shown in Figure 13. Due to the compactness of the compute-intensive units of ELSA, which are on the critical path, ELSA can run 3.2X faster in terms of clock frequency. While the ratio of the energy efficiency is maintained at 1.2X moving from Figure 12 to Figure 13, the area efficiency of ELSA is greatly improved and reaches 3.6X. This is mainly due to the increase in the peak performance as the increase in the cell area was negligible and the ratio remains at 0.3X. Although running ELSA at its highest clock frequency increased its power consumption, it is still lower (0.9X) than that of the Baseline-LSTM.

6.2.2 **Comparison with the Existing ASIC Implementations**. ELSA is also compared against the existing ASIC implementations of LSTMs – DNPU [27] and CHIPMUNK [5] as shown in Table 4. DNPU is a CNN-RNN processor and its application requires a combination of CNNs and RNNs. CNN is its major component and RNN was not evaluated as a standalone component. Although ELSA has twice the bit-precision and uses 10X more SRAMs than DNPU, it achieves higher peak-performance and consumes less power. DNPU's bit-width (4-bits) is half of ELSA's. Scaling ELSA to 4-bits would increase the peak-performance (at-least 54 GOPs) and the frequency (~400MHz), and decrease the power consumption. These would lead to a much higher energy-efficient design. In addition, DNPU has only 10KB of on-chip memory which limits its peak-performance by requiring the use of external memory even for small networks. The application in which the functionality of ELSA was evaluated on (even for 4-bits), does not fit on DNPU and requires a DRAM. This lowers DNPU's peak-performance substantially. CHIPMUNK uses 22% smaller SRAMs. It achieves higher peak-performance, but it consumes 30% more power, making ELSA more energy-efficient. As shown in the last entry of Table 4, ELSA's energy-efficiency exceeds that of DNPU and CHIPMUNK by 1.2X and 1.18X, respectively.

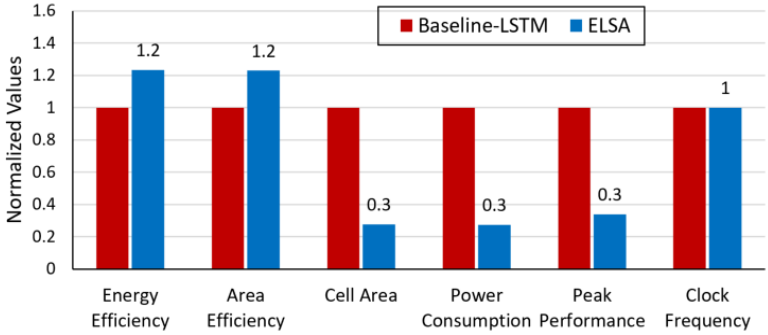

Fig. 12. The ASIC implementation results of ELSA as compared to the Baseline-LSTM. Both of these designs were run at the same clock frequency, the highest that the Baseline-LSTM can achieve. The reported numbers are normalized. The energy and area efficiency of ELSA exceeds that of the Baseline-LSTM by factors of 1.2X.

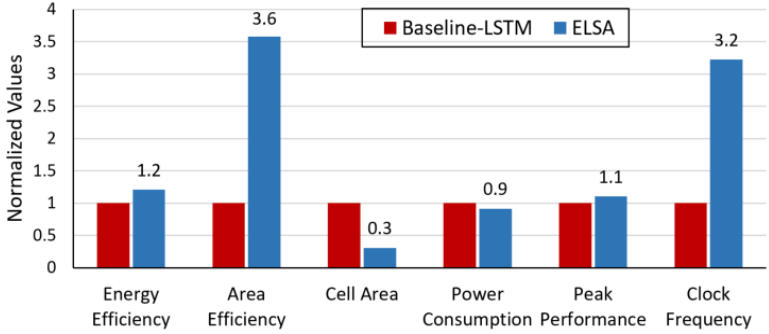

Fig. 13. The ASIC implementation results of ELSA as compared to the Baseline-LSTM. Both of these designs were run at their highest achievable clock frequency. The reported numbers are normalized. The energy and area efficiency of ELSA exceeds that of the Baseline-LSTM by factors of 1.2X and 3.6X, respectively.

### 6.3 Accuracy versus Hardware Bit Precision

This section describes two main explorations on the accuracy of ELSA. First, it examines the impact of using the AMs on error propagation through the LSTM for different bit precision. Specifically, it investigates whether the error accumulates in the hidden and memory states over various time-steps. The baseline design against which the precision of ELSA is compared, is a software implementation using floating point calculations. The precision of ELSA is also compared with a design that uses exact fixed-point multiplications.

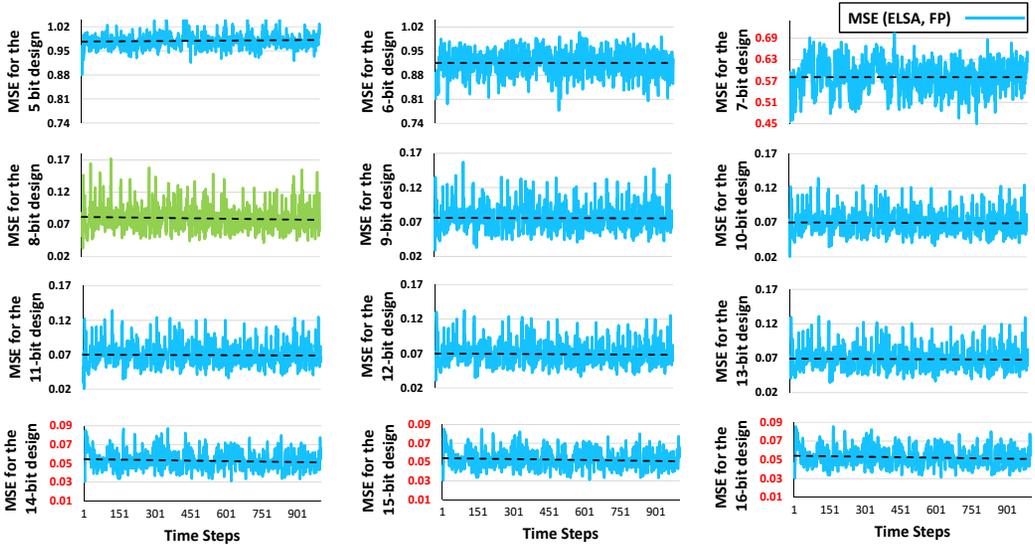

Fig. 14. The MSE of the hidden state of ELSA and the floating-point (FP) for different bit-precision. Each plot depicts the MSE for a specific precision over 1000 time steps. The dashed black line shows that the error does not accumulate at the application level. The MSE decreases substantially as the bit precision increases from 5 to 8 bits. From the 8-bit design and above, the magnitude of MSE is close to zero and it does not decrease significantly with the increase in the precision. The same trend is true for the memory state, for which the plot is not included for brevity. The 8-bit design is a good choice for LM as the MSE is very close to zero and the decreasing rate becomes smaller from 8 to 16 bits.

Table 4. Comparison with the previous ASIC implementations. All of these implementations are in 65nm technology. DNPU is a CNN-RNN processor, and this table only includes the RNN values reported in [27]. The LSTM architecture of DNPU and CHIPMUNK differ substantially among themselves and also when compared with ELSA. Moreover, the reported applications are dramatically different, making comparisons in general difficult to judge.

|  | DNPU [27] | CHIPMUNK [5] | ELSA |
|---|---|---|---|
| **Precision (bit)** | 4-7 | 8-16 | 8-11 |
| **Frequency (MHz)** | 200 | 168 | 322 |
| **On-Chip Memory (KB)** | 10 | 82 | 106 |
| **Peak Performance (GOP/s)** | 25 | 32.3 | 27 |
| **Power (mW)** | 21 | 29.03 | 20.4 |
| **Energy-Efficiency (GOP/s/mW)** | 1.10 | 1.11 | **1.32** |
| **ELSA's Energy-Efficiency (X)** | **1.2** | **1.18** | 1 |

### 6.3.1 *Comparison with software floating-point implementation.*
Due to the recurrent nature of the LSTM on the memory and hidden states, a thorough comparison of the accuracy is performed at an application level (i.e. LM) for both $h$ and $C$. For a fair comparison, the same input sequence $X_t$ was fed to both designs, for $t \in [1, 2, ..., 1000]$. The accuracy was computed as the mean squared error (MSE) between ELSA's results and the floating-point (FP) implementation, for which the results are displayed in Figure 14 for various bit-width for ELSA. The black line on each plot is the trend-line across all the time-steps. The key observation here is that the error does not accumulate and does not grow – a behavior that is consistent for all the bit precisions. This is because of the inherent feature of the AM that rounds up/down the final product based on the given inputs. This has the effect of canceling the errors. The MSE trend line is very close to zero at 8-bits, and remains nearly the same up to 16 bits. This is due to the good accuracy of the AM for 8 bits and above. The same trend is true for the memory state of ELSA, for which the plots are omitted for brevity. Based on these experiments, to achieve a good trade-off between the accuracy of ELSA and its hardware design metrics (i.e. power, area, performance), the bit-precision was set to 8 bits.

*6.3.2 Comparison with Exact fixed-point implementation.* The accuracy of ELSA was compared to an LSTM design with exact fixed-point multipliers, assuming a bit precision of 8 for both. This experiment demonstrates how the accuracy changes from a single AM up to a network of LSTMs. Table 5 shows the accuracy for a single multiplication, a MAC unit, an LSTM layer and an application (i.e. LM that has two consecutive LSTM layers), when the AM is employed. The interpretation of Table 5 is as follows.

(1) The accuracy for one multiplication and MAC operation was computed as the fraction of differences between the 8-bit AM and its corresponding 8-bit fixed point exact multiplier. That is, for each input, the relative error of every pair of corresponding multipliers and MAC units was computed and these values were averaged over the set of applied inputs. The mean and standard deviation (Std. Dev.) are also reported.
(2) The accuracy for LSTM (ELSA) was measured as the average accuracy of the hidden states. This is the relative error between the corresponding values of $h_t$ (see Figure 8) in the 8-bit AM and 8-bit fixed point exact multiplier, averaged over the set of applied inputs.
(3) The last entry of Table 5 reports the classification accuracy of ELSA at the application level. The accuracy degraded by 2.5% when moving from an AM to a full application which consists of a network of consecutive LSTMs. The Top-5 classification accuracy (a standard measure particularly for LM) was 96%.

Table 5. The accuracy for a single multiplication, MAC operations, an LSTM layer and an application (i.e. LM that has two consecutive LSTM layers), when the AM is employed for the 8-bit hardware bit precision. Each AM-based unit is compared against its corresponding 8-bit exact fixed-point unit.

|  | Relative Error | Mean of Relative Error | Std. Dev. of Relative Error |
| --- | --- | --- | --- |
| One Multiplication | 1.5% | 0.00562 | 0.00415 |
| MAC Operations | 2.2% | 0.00566 | 0.00416 |
| LSTM (one layer) | 2.3% | 0.00181 | 0.00149 |

|  | **Classification Accuracy** |
| --- | --- |
| Application (LM) | **96%** |

## 7 CONCLUSIONS AND FUTURE WORK

This paper presents a novel scalable LSTM hardware accelerator, referred to as ELSA, that results in small area and high energy-efficiency. This is due to several architectural features, including the use of an improved low-power, compact approximate multiplier in the compute-intensive units of ELSA, and the design of two levels of controllers that are required for handling the variable-cycle multiplications. Moreover, ELSA includes efficient synchronization of the elastic pipeline stages to maximize the utilization. ELSA achieves promising results in power, area and energy-efficiency making it suitable for use in embedded systems and real-time applications. This accelerator can be further improved by incorporating more compact SRAMs to achieve a more optimized floor-plan. In addition, the energy-efficiency can be significantly improved by applying weight compression techniques.

## ACKNOWLEDGMENTS

This work was supported by the NSF I/UCRC Center for Embedded Systems and NSF grant number 1361926.